\documentclass{article} %
\usepackage{iclr2026_conference,times}

\usepackage{amsmath,amsfonts,bm}

\def\eqref#1{equation~\ref{#1}}

\def\1{\bm{1}}

\DeclareMathAlphabet{\mathsfit}{\encodingdefault}{\sfdefault}{m}{sl}
\SetMathAlphabet{\mathsfit}{bold}{\encodingdefault}{\sfdefault}{bx}{n}

\usepackage{booktabs}   %
\usepackage{multirow}   %
\usepackage{graphicx}   %
\usepackage[utf8]{inputenc} %
\usepackage[T1]{fontenc}    %
\usepackage{url}            %
\usepackage{amsfonts}       %
\usepackage{nicefrac}       %
\usepackage{microtype}      %
\usepackage{xcolor}         %
\usepackage{xspace}
\usepackage{wrapfig}
\usepackage{hyperref}
\usepackage[capitalize]{cleveref}
\usepackage{natbib}       %
\usepackage{amsmath}
\usepackage{bm}
\usepackage{wrapfig} %
\usepackage{amssymb,mathtools}

\renewcommand{\vec}[1]{\bm{#1}}
\newcommand{\mat}[1]{\mathbf{#1}}

\def\modelname{Durian}
\def\prnet{PRNet}
\def\arnet{ARNet}
\def\dnet{DNet}

\def\paperTitle{\modelname: Dual Reference Image-Guided \\ Portrait Animation with Attribute Transfer}

\definecolor{tabfirst}{rgb}{1, 0.7, 0.7} %
\definecolor{tabsecond}{rgb}{1, 0.85, 0.7} %
\definecolor{tabthird}{rgb}{1, 1, 0.7} %

\usepackage{soul}      %
\usepackage{xcolor}    %
\sethlcolor{yellow}    %

\title{\paperTitle}

\author{Hyunsoo Cha, Byungjun Kim, Hanbyul Joo\\
Seoul National University\\
\texttt{\{243stephen,byungjun.kim,hbjoo\}@snu.ac.kr}
}

\iclrfinalcopy %

\begin{document}

\maketitle

\begin{figure}[h]
  \includegraphics[width=\textwidth]{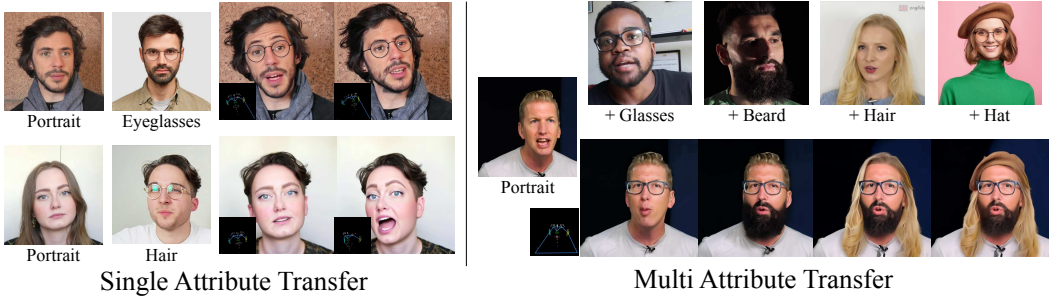}\centering
  \vspace{-1em}
  \caption{
    \textbf{Portrait Animation with Attribute Transfer.} 
    Given a portrait image and single or multiple reference images specifying target attributes (\emph{e.g.}, hairstyle, eyeglasses), our method generates a portrait animation with facial attribute transfer conditioned on a keypoint sequence.
  }
  \label{fig:teaser}
\end{figure}

\begin{abstract}
We present \modelname, the first method for generating portrait animation videos with cross-identity attribute transfer from one or more reference images to a target portrait.
Training such models typically requires attribute pairs of the same individual, which are rarely available at scale.
To address this challenge, we propose a self-reconstruction formulation that leverages ordinary portrait videos to learn attribute transfer without explicit paired data.
Two frames from the same video act as a pseudo pair: one serves as an attribute reference and the other as an identity reference.
To enable this self-reconstruction training, we introduce a Dual ReferenceNet that processes the two references separately and then fuses their features via spatial attention within a diffusion model.
To make sure each reference functions as a specialized stream for either identity or attribute information, we apply complementary masking to the reference images.
Together, these two components guide the model to reconstruct the original video, naturally learning cross-identity attribute transfer.
To bridge the gap between self-reconstruction training and cross-identity inference, we introduce a mask expansion strategy and augmentation schemes, enabling robust transfer of attributes with varying spatial extent and misalignment.
\modelname{} achieves state-of-the-art performance on portrait animation with attribute transfer.
Moreover, its dual reference design uniquely supports multi-attribute composition and smooth attribute interpolation within a single generation pass, 
enabling highly flexible and controllable synthesis. 
Project page and code are available at: \color{magenta}{\url{https://hyunsoocha.github.io/durian}}
\end{abstract}
\vspace{-1em}

\section{Introduction}
\vspace{-0.7em}

Personalized appearance editing, such as virtually trying on glasses or experimenting with new hairstyles, is becoming a key feature of virtual styling applications.
However, most existing solutions are highly specialized and limited in scope.
Hairstyle preview apps typically rely on fixed templates, which may look realistic from a single view but fail to adapt to head pose or expression changes. %
Glasses try-on systems often depend on pre-scanned 3D product models, restricting users to a predefined catalog. %
Furthermore, these systems focus on a single attribute and cannot combine multiple elements, such as hair, glasses, or hats, within a unified experience.

A key challenge in building such a system is obtaining suitable training data.
Disentangling identity from attributes ideally requires paired images of the same person with different attributes, which are rarely available and expensive to collect at scale.
This difficulty grows exponentially for multiple attributes, as capturing all combinations quickly becomes infeasible.
For example, \citet{li2023megane} collect multi-view images of subjects wearing different eyeglasses to model realistic glasses try-on, but the dataset remains too limited to generalize broadly.
\citet{zhang2025stablehair} propose a synthetic pipeline that predicts a bald version of a portrait and generates reference hair images using a pretrained diffusion model.
However, this approach is not easily scalable beyond hair.

This naturally raises the question: \emph{can we train a model for portrait animation with attribute transfer without any explicit attribute-paired data?}
Motivated by this question, we propose a \textbf{self-reconstruction framework} that learns this task directly from widely available in-the-wild portrait videos.
During training, we randomly sample two frames from a single video: one as the attribute reference and the other as the identity reference. 
The remaining frames are treated as targets to be generated, conditioned on a keypoint sequence representing the motion of the video.
To prevent identity leakage, we apply complementary masking to the two reference frames so that the network must disentangle and combine the attribute and identity information to reconstruct the original video.

To enable this framework, we design a \textbf{Dual ReferenceNet} architecture that explicitly encodes the attribute and portrait references through two separate branches and fuses their disentangled features for generation via spatial attention.
This design enables the network to move beyond simple pose driving, generating keypoint-driven portrait animations that seamlessly combine the attribute from one image with the identity from the other.
Surprisingly, although the model is trained with only a single attribute reference at a time, the spatial attention mechanism allows more advanced operations at inference time.
Since different attributes (\emph{e.g.}, hair, glasses, beard, hats) occupy distinct spatial regions, their features can be jointly injected without conflict, enabling seamless multi-attribute transfer.
Furthermore, by interpolating the features of two attribute references, our model can achieve attribute interpolation, generating smooth transitions between the attributes.
These emergent capabilities make our framework especially valuable for real-world styling scenarios, where users may want to explore diverse combinations and gradual transformations of facial attributes.

While self-reconstruction training is effective for learning to separate identity and attributes, it operates within a single video, leading to a domain gap when the model is applied to cross-identity inference, where the attribute and portrait come from different individuals.
To mitigate this gap, we introduce a mask expansion strategy and lightweight augmentation schemes.
These techniques expose the model to a broader range of attribute configurations during training, enabling robust transfer across spatial and structural variations of the attribute region.
These designs form a unified framework capable of robust cross-identity attribute transfer.
As a result, our method achieves a versatile system that generates portrait animations with diverse appearance edits in a zero-shot manner.

We summarize the key contributions of our work, as follows: 
\begin{itemize}
    \item We propose the first method to generate keypoint-driven portrait animations with transferred attributes directly from two images, generalized across diverse facial attributes beyond hair.
    \item We design a Dual ReferenceNet architecture that disentangles attribute and identity through two branches fused via spatial attention, enabling self-reconstruction training directly on uncurated in-the-wild videos without paired data.
    \item We propose a mask expansion strategy and lightweight augmentations to bridge the domain gap for cross-identity transfer, improving robustness to diverse spatial configurations.
    \item Our framework exhibits an emergent ability to support multi-attribute composition and interpolation in a single generation pass, without requiring any additional training.
\end{itemize}

\section{Related Work}
\paragraph{Face Editing.}
Generative models have advanced facial editing from unconditional synthesis to fine-grained manipulation of existing images~\citep{goodfellow2014gan,rezende2015normalizingflow,ho2020ddpm}. Latent-space editing with StyleGAN~\citep{karras2020stylegan2} and GAN inversion~\citep{zhu2016ganinversion,abdal2019image2stylegan,richardson2021encoding} has been extended to video via latent trajectory modeling~\citep{yao2021latenttransformer,tzaban2022stitch} and 3D-aware editing~\citep{bilecen2024reference,xu2024innout}. However, such approaches often rely on attribute classifiers or fixed editing controls. Diffusion-based models have introduced more flexible editing through prompt-driven~\citep{brooks2023instructpix2pix} or identity-preserving techniques~\citep{ye2023ipadapter,wang2024instantid}, with extensions to video improving temporal consistency~\citep{ku2024anyv2v,kim2023diffusionvideoae}. Still, these methods are limited to modifying existing content and cannot generate new motions or expressions.

\paragraph{Diffusion-based Attribute Transfer.}
Diffusion-based attribute transfer methods typically formulate editing as masked inpainting, where reference content is inserted into a target image using explicit masks~\citep{yang2023pbe, chen2024anydoor, mou2025dreamo, chen2025edittransfer, song2025insertanything}. 
These approaches have been adapted to domain-specific tasks such as hairstyle~\citep{zhang2025stablehair, chung2025hairfusion}, clothing~\citep{kim2024stableviton, li2024anyfit, chong2024catvton}, and makeup~\citep{zhang2024stablemakeup}. While effective for static images, they rely on category labels or mask annotations. Video extensions~\citep{fang2024vivid, tu2025videoanydoor} apply per-frame inpainting with post-hoc smoothing, but predefined masks are hard to specify for deformable facial attributes that vary over time. Recent works have also explored attribute transfer in 3D avatars~\citep{kim2024gala,nam2025decloth,cha2024pegasus,cha2025perse,wang2025mega,kim2025haircup}, but such approaches often require specialized capture setups or are not easily generalizable to in-the-wild scenarios.
In contrast, our model performs attribute transfer and animation jointly in a single forward pass, conditioned only on a pair of reference images and a facial keypoint sequence. This eliminates the need for per-frame masks, text prompts, or category labels, enabling zero-shot transfer of diverse facial attributes.

\paragraph{Portrait Animation from a Single Image.}
Portrait animation aims to generate motion from a static image, typically guided by facial keypoints, audio, or motion trajectories. Early methods rely on GANs with implicit keypoint modeling~\citep{guo2024liveportrait,wang2021facevid2vid}, while recent approaches use diffusion models~\citep{hu2024animate,zhu2024champ,yang2025megactor} for improved realism and temporal stability. These methods primarily focus on reenactment and identity preservation. Others incorporate paired motion~\citep{xie2024x} or audio~\citep{yang2025megactor}, but require multi-stage inference or fine-tuning.
Our model jointly performs facial attribute transfer and motion generation, producing photorealistic, identity-preserving videos from diverse attribute references and keypoint-driven motion in a single pass.
\begin{figure*}[t]
  \centering
  \includegraphics[width=\textwidth]{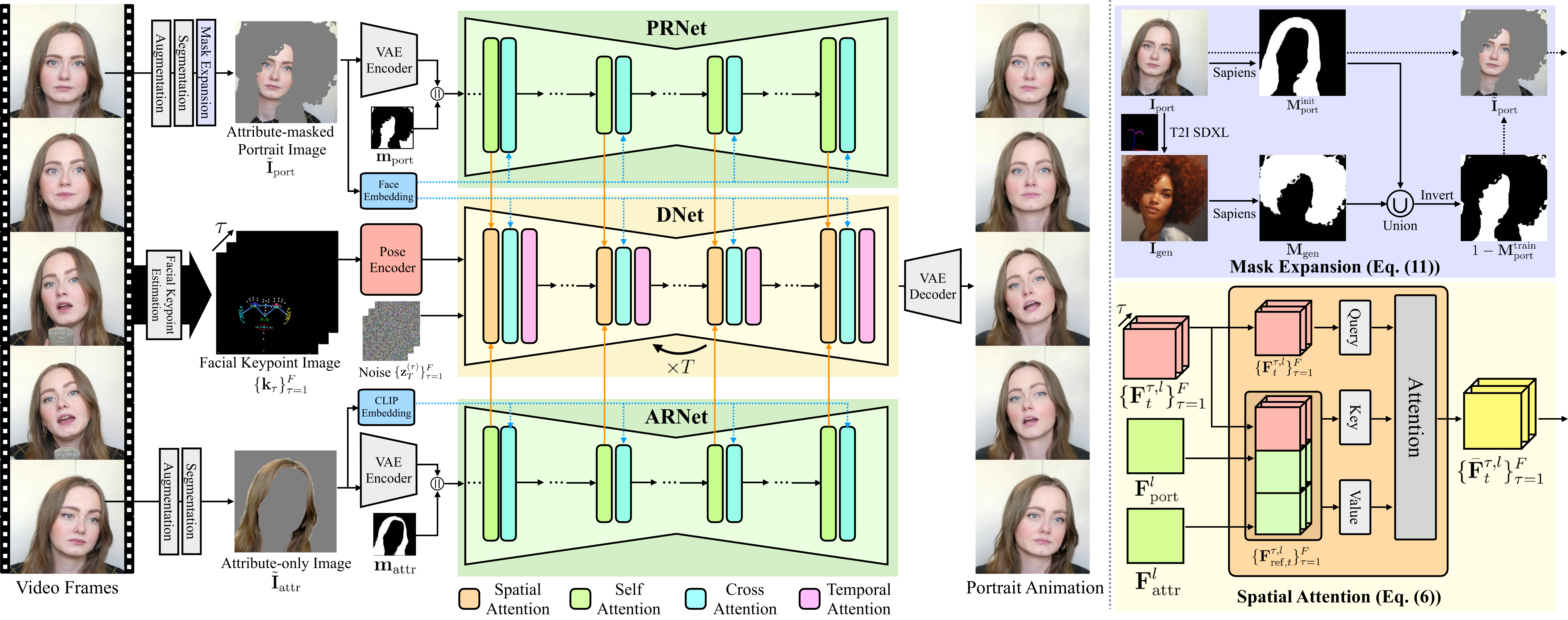}
    \caption{
    \textbf{Overview of Training Pipeline.}
    Given an attribute-masked portrait image $\tilde{\mat{I}}_\mathrm{port}$ and an attribute-only image $\tilde{\mat{I}}_\mathrm{attr}$, \modelname{} synthesizes a portrait animation with the transferred attribute. These inputs are constructed by randomly sampling two frames from a training video and applying the estimated masks. A sequence of facial keypoints $\{{\vec{k}_\tau}\}_{\tau = 1}^{F}$ is extracted from the video to guide the motion. During generation, spatial features from \prnet{} and \arnet{} are fused via spatial attention into the \dnet, ensuring identity preservation and attribute consistency in the synthesized video.
    }
  \label{fig:model_overview}
\end{figure*}

\section{Method}

\subsection{Overview: Learning Attribute Transfer from Self-Reconstruction}
\label{sec:overview}
We propose a diffusion-based generative framework for portrait animation with cross-identity attribute transfer. 
At a high level, our model generates an $F$-frame animation sequence $\mat{V}=\{\mat{I}_{\tau}\}_{\tau=1}^{F}$ as:

\begin{equation}
\mathbf{V} = \mathrm{\modelname}(\mat{I}_{\mathrm{attr}}, \mat{M}_{\mathrm{attr}},
                                 \mat{I}_{\mathrm{port}}, \mat{M}_{\mathrm{port}}, 
                                 \mat{K}),
\end{equation}

conditioned on an attribute image $\mat{I}_{\mathrm{attr}}$, a portrait image $\mat{I}_{\mathrm{port}}$, and a sequence of driving facial keypoint images $\mat{K} = \{\vec{k}_\tau\}_{\tau=1}^F$.
Each reference image has a binary mask: $\mat{M}_{\mathrm{attr}}$ localizes the attribute region (\emph{e.g.}, hair or glasses) in the reference image, while $\mat{M}_{\mathrm{port}}$ specifies the candidate region in the portrait where the attribute will be transferred.
Using these masks, we construct two masked inputs: the \emph{attribute-only image} $\tilde{\mat{I}}_{\mathrm{attr}} = \mat{I}_{\mathrm{attr}} \odot \mat{M}_{\mathrm{attr}}$, where only the attribute region is preserved, and the \emph{attribute-masked portrait image} $\tilde{\mat{I}}_{\mathrm{port}} = \mat{I}_{\mathrm{port}} \odot (1 - \mat{M}_{\mathrm{port}})$, where the corresponding region is removed.
These masked inputs are fed into the \textbf{Dual ReferenceNet}, consisting of the \emph{Attribute ReferenceNet (ARNet)} and \emph{Portrait ReferenceNet (PRNet)}, which extract multi-scale spatial features. These features are then injected into a diffusion-based generator, the \emph{Denoising UNet (DNet)}, to synthesize the remaining frames of the video with keypoint guidance $\mat{K}$ (\cref{sec:architecture}).

To enable training without requiring explicitly annotated triplets (\emph{i.e.}, combinations of a target attribute image, an original portrait image, and an edited portrait image), we adopt a \textbf{self-reconstruction strategy} based on portrait videos~\citep{yu2023celebv, xie2022vfhq}. Specifically, we simulate attribute transfer by sampling two frames $\mat{I}_{\mathrm{attr}}$ and $\mat{I}_{\mathrm{port}}$ from the same video, treating one as the attribute reference and the other as the target portrait.
We then construct the masked inputs $\tilde{\mat{I}}_{\mathrm{attr}}$ and $\tilde{\mat{I}}_{\mathrm{port}}$ using the same masking formulation as in inference, based on a segmentation mask of a randomly selected attribute.
Although the two frames come from the same identity, the complementary masking enforces a clear separation between identity and attribute inputs, encouraging the model to learn meaningful mappings from these features to output frames without requiring cross-identity supervision.
To enhance the model’s ability to generalize beyond the self-attribute transfer setup, we introduce an augmentation scheme that improves robustness to spatial and appearance variations(\cref{sec:training}). 

At inference time, we estimate refined attribute masks by aligning the attribute image to the portrait through a lightweight alignment process, mitigating spatial misalignment between them.
Conditioned on the two masked reference images and the driving keypoint sequence, our model then synthesizes portrait animations with attribute transfer.
Notably, our design also supports multi-attribute composition and smooth interpolation within a single generation pass, without requiring additional training or post-processing (\cref{sec:inference}).
\cref{fig:teaser} shows our generated portrait animations with attribute transfer.

\subsection{Model Architecture: Dual ReferenceNet}
\label{sec:architecture}
Inspired by recent approaches~\citep{guo2023animatediff, hu2024animate, zhu2024champ} that leverage ReferenceNet to inject spatial features into diffusion models, we propose a \textbf{Dual ReferenceNet} architecture tailored for portrait animation with attribute transfer. Unlike previous work, our model includes two separate encoders: \emph{Attribute ReferenceNet (\arnet{})} and \emph{Portrait ReferenceNet (\prnet{})}, each sharing the same architecture as the \emph{Denoising U-Net (\dnet{})} in the diffusion model, excluding the temporal layers. The networks follow the U-Net~\citep{long2015unet} architecture used in latent diffusion models~\citep{rombach2022ldm}, with each block containing convolutional layers followed by self- and cross-attention modules. The overall architecture is shown in \cref{fig:model_overview}.

\paragraph{Reference inputs.}
Given an attribute image $\mat{I}_{\mathrm{attr}}\in \mathbb{R}^{3 \times H \times W}$ and a portrait image $\mat{I}_{\mathrm{port}}\in \mathbb{R}^{3 \times H \times W}$, along with their binary masks $\mat{M}_{\mathrm{attr}}\in \mathbb{R}^{1 \times H \times W}$ and $\mat{M}_{\mathrm{port}}\in \mathbb{R}^{1 \times H \times W}$, which localize the attribute region and the candidate transfer region respectively, we construct two masked inputs:
the attribute-only image $\tilde{\mat{I}}_{\mathrm{attr}} = \mat{I}_{\mathrm{attr}} \odot \mat{M}_{\mathrm{attr}}$, where only the attribute region is preserved, and the attribute-masked portrait image $\tilde{\mat{I}}_{\mathrm{port}} = \mat{I}_{\mathrm{port}} \odot (1 - \mat{M}_{\mathrm{port}})$, where the corresponding candidate region is removed.
We then encode these masked images into latent representations using the pretrained VAE from the latent diffusion model~\citep{rombach2022ldm}, yielding $\vec{z}_{\mathrm{attr}}, \vec{z}_{\mathrm{port}} \in \mathbb{R}^{c \times h \times w}$.
The corresponding masks $\mat{M}_{\mathrm{attr}}, \mat{M}_{\mathrm{port}}$ are downsampled to match the latent resolution, producing $\vec{m}_{\mathrm{attr}}, \vec{m}_{\mathrm{port}} \in \mathbb{R}^{1 \times h \times w}$. These downsampled masks are concatenated with the latents along the channel dimension to form $(c+1)$-channel inputs $\tilde{\vec{z}}_{\mathrm{attr}}, \tilde{\vec{z}}_{\mathrm{port}} \in \mathbb{R}^{(c+1) \times h \times w}$ as follows:

\begin{equation}
\tilde{\vec{z}}_{\mathrm{attr}}
= \mathrm{concat}_{\mathrm{c}}(\vec{z}_{\mathrm{attr}}, \vec{m}_{\mathrm{attr}}), \quad
\tilde{\vec{z}}_{\mathrm{port}}
= \mathrm{concat}_{\mathrm{c}}(\vec{z}_{\mathrm{port}}, \vec{m}_{\mathrm{port}}).
\end{equation}

\paragraph{Spatial attention.}
The augmented latents are passed to \arnet{} $\mathcal{E}_{\mathrm{attr}}$ and \prnet{} $\mathcal{E}_{\mathrm{port}}$ to extract multi-scale feature maps after convolutional layers of each block:
\begin{equation}
\label{eq:ref_feat}
\mathcal{F}_{\mathrm{attr}} \coloneq \{ \mat{F}_{\mathrm{attr}}^{l} \}_{l=1}^{L} 
= \mathcal{E}_{\mathrm{attr}}(\tilde{z}_{\mathrm{attr}}; \Theta_{\mathrm{attr}}), \quad
\mathcal{F}_{\mathrm{port}} \coloneq \{ \mat{F}_{\mathrm{port}}^{l} \}_{l=1}^{L} 
= \mathcal{E}_{\mathrm{port}}(\tilde{z}_{\mathrm{port}}; \Theta_{\mathrm{port}}),
\end{equation}
where $\Theta_{\{\mathrm{attr,port}\}}$ are the parameters of Dual ReferenceNet.
Let $\mat{F}_t^{\tau, l} \in \mathbb{R}^{c_l \times h_l \times w_l}$ denote the feature map of the frame $\tau$ at the $l$-th block of the denoising U-Net.
While the original denoising U-Net includes a self-attention layer at each resolution, we replace it with our spatial attention to integrate identity and attribute features in a spatially-aware manner. 
We denote width-wise concatenation as $\mathrm{concat}_{\mathrm{w}}(\cdot)$, and define our spatial attention $\mathrm{SA}(\cdot, \cdot, \cdot)$ as:

\begin{equation}
\label{eq:concat_def}
\mat{F}_{\mathrm{ref}, t}^{\tau, l} \coloneq \mathrm{concat}_{\mathrm{w}}(\{\mat{F}_t^{\tau, l}, \mat{F}_{\mathrm{port}}^l, \mat{F}_{\mathrm{attr}}^l\}) \in \mathbb{R}^{c_l \times h_l \times 3w_l},
\end{equation}
\begin{equation}
\label{eq:sa_def}
\bar{\mat{F}}_t^{\tau,l}=\mathrm{SA}(
\mat{F}_t^{\tau, l}, 
\mat{F}_{\mathrm{port}}^l, 
\mat{F}_{\mathrm{attr}}^l)
= \mathrm{Attention}(\mat{W}_Q \mat{F}_t^{\tau, l}, \mat{W}_K \mat{F}_{\mathrm{ref}, t}^{\tau, l}, 
\mat{W}_V \mat{F}_{\mathrm{ref}, t}^{\tau, l}),
\end{equation}

where $\bar{\mat{F}}_t^{\tau, l}\in \mathbb{R}^{c_l \times h_l \times w_l}$ is the feature map after the spatial attention, $\mathrm{Attention}(Q, K, V) = \mathrm{softmax}({QK^\top} / {\sqrt{d}})V$
is the standard scaled dot-product attention~\citep{vaswani2017transformer}, $\mat{W}_Q, \mat{W}_K, \mat{W}_V$ are linear projection layers.
This width-wise concatenation preserves spatial resolution and allows the model to attend across all positions in the combined reference and target features. 
As a result, the model can leverage both attribute and portrait guidance at every step.

\paragraph{Cross-attention with semantic embeddings.}
After applying spatial attention, we further inject semantic guidance into both the Dual ReferenceNet and the denoising U-Net via cross-attention.
For \arnet, we use the CLIP~\citep{radford2021learning} embedding of the attribute-only image $\tilde{\mat{I}}_{\mathrm{attr}}$ as the attribute embedding $\vec{\phi}_{\mathrm{attr}}$, which is injected via cross-attention into each block of \arnet.
For \prnet{} and \dnet, we construct a portrait embedding $\vec{\phi}_{\mathrm{port}}$ by combining ArcFace~\citep{deng2019arcface} and CLIP embeddings of the attribute-masked portrait image $\tilde{\mat{I}}_{\mathrm{port}}$ following StableAnimator~\citep{tu2024stableanimator}. This embedding is injected into both \prnet{} and \dnet{} to enhance identity preservation.
We define the cross-attention operation $\mathrm{CA}(\cdot, \cdot)$ as:
\begin{equation}
\mathrm{CA}(\bar{\mat{F}},\, \vec{\phi}) = 
\mathrm{Attention}(\mat{W}'_Q \bar{\mat{F}},\, \mat{W}'_K \vec{\phi},\, \mat{W}'_V \vec{\phi}),
\end{equation}
where $\bar{\mat{F}}$ is the input feature map, $\vec{\phi}$ is the conditioning embedding, and $\mat{W}'_Q, \mat{W}'_K, \mat{W}'_V$ are learned linear projections.
Let $\bar{\mat{F}}_{\mathrm{attr}}^l$ and $\bar{\mat{F}}_{\mathrm{port}}^l$ be the self-attended features of the $l$-th block in \arnet{} and \prnet{}, and $\bar{\mat{F}}_t^l$ the spatially attended feature of \dnet{}.
Then, the cross-attention updates are given by:
\begin{equation}
\label{eq:ca_def}
\tilde{\mat{F}}_{\{\mathrm{attr, port}\}}^{l} = \mathrm{CA}(\bar{\mat{F}}_{\{\mathrm{attr, port}\}}^{l},\, \vec{\phi}_{\{\mathrm{attr, port}\}}), \quad
\tilde{\mat{F}}_t^{\tau, l} = \mathrm{CA}(\bar{\mat{F}}_t^{\tau, l},\, \vec{\phi}_{\mathrm{port}}),
\end{equation}
where $\tilde{\mat{F}}_{\mathrm{attr}}^{l}$, $\tilde{\mat{F}}_{\mathrm{port}}^{l}$, and $\tilde{\mat{F}}_t^{\tau, l}$ are the feature maps after cross-attention in \arnet, \prnet, and \dnet.

\paragraph{Temporal extension and keypoint guidance.}
Our model incorporates temporal awareness to generate coherent portrait animations by inserting temporal self-attention into each U-Net block, following~\citet{hu2024animate, zhu2024champ}.
To control pose and expression, we use a sequence of facial keypoints $\mat{K} = \{\vec{k}_\tau\}_{\tau=1}^{F}$ extracted by Sapiens~\citep{khirodkar2024sapiens}. Each keypoint image $\vec{k}_\tau$ is encoded into a spatial feature map $\mat{F}_{\mathrm{kpt}}^\tau$ via a pose encoder and combined with the noisy latent $\vec{z}_t^{(\tau)}$ following~\citet{zhu2024champ}.
For each frame $\tau$, \dnet{} $\epsilon_\theta$ predicts the added noise $\hat{\epsilon}^{(\tau)}_t$ from the noisy latent $\vec{z}_t^{(\tau)}$ at timestep~$t$, using the reference features, semantic embeddings, and keypoint features:
\begin{equation}
\label{eq:denoise}
\hat{\epsilon}^{(\tau)}_t = \epsilon_\theta\left(
\vec{z}_t^{(\tau)},\, t,\,
\mathcal{F}_{\mathrm{attr}},
\mathcal{F}_{\mathrm{port}},
\vec{\phi}_{\mathrm{attr}},
\vec{\phi}_{\mathrm{port}},
\mat{F}_{\mathrm{kpt}}^\tau
\right).
\end{equation}
The predicted noise is used to recover the denoised latent $\vec{z}_0^{(\tau)}$, then decoded by the VAE decoder $\mathcal{D}$ to produce the final video frame as $\mat{I}_\tau = \mathcal{D}(\vec{z}_0^{(\tau)})$ for $\tau = 1, \dots, F$.

\subsection{Training Strategy}
\label{sec:training}
\paragraph{Training loss.}To effectively train our model, we adopt a two-stage training scheme following the previous approaches~\citep{hu2024animate,zhu2024champ}. 
In the first stage, we optimize the entire model except the temporal attention layers, treating each video frame as an independent training sample.
We define the per-frame conditioning bundle as \(
\mathcal{C} := \left(
\mathcal{F}_{\mathrm{attr}},
\mathcal{F}_{\mathrm{port}},
\vec{\phi}_{\mathrm{attr}},
\vec{\phi}_{\mathrm{port}}
\right),
\)
where $\mathcal{F}_{\mathrm{port}}, \mathcal{F}_{\mathrm{attr}}$ are the multi-scale spatial features from PRNet and ARNet and $\vec{\phi}_{\mathrm{port}}, \vec{\phi}_{\mathrm{attr}}$ are the semantic embeddings. 
Then, the training objective is the standard denoising diffusion loss:
\begin{equation}
\mathcal{L}_{\mathrm{diff}}^{(1)} =
\mathbb{E}_{\vec{z}_0,\, \epsilon,\, t} \left[
\left\| \epsilon - \epsilon_\theta\left(
\vec{z}_t,\, t,\, \mathcal{C},\, \mat{F}_{\mathrm{kpt}}
\right) \right\|^2
\right],
\end{equation}
where $\vec{z}_t$ is the noised latent at diffusion timestep $t$, $\epsilon$ is the sampled noise, and $\mat{F}_{\mathrm{kpt}}$ is the feature map of the corresponding facial keypoint image.
In the second stage, we freeze all modules except the temporal attention layers and train them using multi-frame inputs. The temporal objective considers a sequence of noised latents and corresponding keypoints:
\begin{equation}
\mathcal{L}_{\mathrm{diff}}^{(2)} =
\mathbb{E}_{\{\vec{z}_0^{(\tau)}\}_{\tau=1}^{F},\, \vec{\epsilon}^{1:F},\, t} \left[
\left\| \vec{\epsilon}^{1:F} - \epsilon_\theta\left(
\{\vec{z}_t^{(\tau)}\}_{\tau=1}^{F},\, t,\,
\mathcal{C},\,
\{\mat{F}_{\mathrm{kpt}}^\tau\}_{\tau=1}^{F}
\right) \right\|^2
\right],
\end{equation}
where $\vec{\epsilon}^{1:F} = \{ \epsilon^{(\tau)} \}_{\tau=1}^F$ denotes the per-frame noise sequence. This staged training improves convergence and allows the temporal attention module to focus on modeling motion dynamics without disrupting the spatial fidelity learned in the first stage.

\paragraph{Attribute-aware mask expansion.}
\begin{wrapfigure}{r}{0.5\linewidth}
  \centering
  \includegraphics[width=\linewidth]{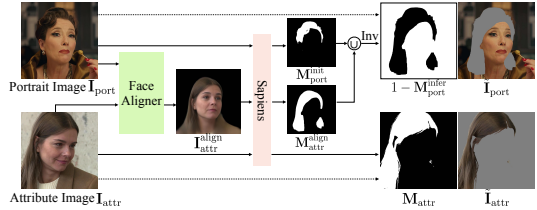}
  \caption{
    \textbf{Aligned Attribute Mask Estimation.}
    To improve attribute-portrait alignment, we estimate an aligned attribute mask via Face Aligner.
  }
  \label{fig:03_attr_mask_est}
\end{wrapfigure}

To expose the model to diverse spatial extents of facial attributes during training, we introduce an attribute-aware mask expansion strategy, illustrated in the top right of \cref{fig:model_overview}.
Given a training frame $\mat{I}$, we first select a target attribute (\emph{e.g.}, hair, eyeglasses, beard) and obtain its binary mask $\mat{M}_{\mathrm{attr}}$ using Sapiens~\citep{khirodkar2024sapiens}.
To simulate variation in the shape and coverage of this attribute, we generate a modified image $\mat{I}_{\mathrm{gen}}$ with SDXL~\citep{podell2023sdxl} and ControlNet~\citep{zhang2023controlnet}, conditioned on the facial keypoints of $\mat{I}$ and a text prompt describing an altered appearance (e.g., “long wavy hair”). 
To enable fully automated prompt generation without any human intervention, we construct a dictionary of descriptive attribute modifiers (e.g., long, short, wavy, curly) and randomly sample their combinations to generate prompts for image generation.
A new mask $\mat{M}_{\mathrm{gen}}$ is then extracted from $\mat{I}_{\mathrm{gen}}$ using Sapiens.
The final training mask is computed as the union of the original and generated masks, and the two masked inputs are constructed as:
\begin{equation}
    \mat{M}_{\mathrm{port}}^{\mathrm{train}} = \mat{M}_{\mathrm{attr}} \cup \mat{M}_{\mathrm{gen}}, \quad
    \tilde{\mat{I}}_{\mathrm{attr}} = \mat{I} \odot \mat{M}_{\mathrm{attr}}, \quad
    \tilde{\mat{I}}_{\mathrm{port}} = \mat{I} \odot (1 - \mat{M}_{\mathrm{port}}^{\mathrm{train}}),
\end{equation}
where $\odot$ denotes element-wise multiplication.
Here, $\mat{M}_{\mathrm{attr}}$ localizes the original attribute region, while $\mat{M}_{\mathrm{port}}^{\mathrm{train}}$ defines the expanded region into which the attribute will be inserted during generation.
This expansion process is \emph{attribute-aware} as it preserves the intended attribute category while diversifying its spatial extent. Unlike HairFusion~\citep{chung2025hairfusion}, which expands masks using fixed heuristics specific to hair, our approach generalizes across multiple facial attributes and enables the model to learn spatially flexible yet semantically grounded transfer patterns.

\paragraph{Reference image augmentation.}
To address the limited diversity of self-reconstruction setups, we introduce an augmentation pipeline that improves robustness to pose, alignment, and appearance variations in attribute–portrait pairs. We perturb both the attribute-only and masked portrait images to simulate realistic spatial and photometric variations. We apply random affine transformations (translation, scaling, rotation) to induce spatial misalignment, and use the FLUX outpainting model~\citep{flux2024} to inpaint newly exposed regions. Additionally, color jittering on tone, contrast, saturation, and hue accounts for appearance variations. This strategy exposes the model to diverse configurations, enabling more robust attribute transfer and animation under real-world variations.

\subsection{Inference Framework and Extensions}
\label{sec:inference}
\paragraph{Inference pipeline.}
At inference time, our system takes as input a portrait image, an attribute image, and a keypoint sequence.  
We first construct two masked reference images: the attribute-only image $\tilde{\mat{I}}_{\mathrm{attr}}$ and the attribute-masked portrait image $\tilde{\mat{I}}_{\mathrm{port}}$, by applying segmentation masks predicted by Sapiens~\citep{khirodkar2024sapiens} to the attribute image $\mat{I}_{\mathrm{attr}}$ and the portrait image $\mat{I}_{\mathrm{port}}$.
To improve spatial alignment between the attribute and portrait inputs, we introduce a \emph{Face Aligner} module, which repurposes a lightweight image-to-3D avatar model~\citep{chu2024gagavatar} solely for alignment.  
This module reconstructs a coarse 3D avatar from the attribute image and aligns its shape and pose to the portrait using FLAME~\citep{li2017flame} parameters $(\vec{\beta}, \vec{\theta}, \vec{\psi})$ estimated by EMOCA~\citep{danvevcek2022emoca}.  
From the resulting pose-aligned image $\mat{I}_{\mathrm{attr}}^{\mathrm{align}}$, we extract a refined attribute mask $\mat{M}_{\mathrm{attr}}^{\mathrm{align}}$ using Sapiens. 
This mask is then merged with the initial portrait mask $\mat{M}_{\mathrm{port}}^{\mathrm{init}}$ to define the final transferable region $\mat{M}_{\mathrm{port}}^{\mathrm{infer}} = \mat{M}_{\mathrm{port}}^{\mathrm{init}} \cup \mat{M}_{\mathrm{attr}}^{\mathrm{align}}$.
The updated mask is applied to construct the final attribute-masked portrait image, 
$\tilde{\mat{I}}_{\mathrm{port}} = \mat{I}_{\mathrm{port}} \odot (1 - \mat{M}_{\mathrm{port}}^{\mathrm{infer}})$,
as illustrated in \cref{fig:03_attr_mask_est}.
Finally, spatial features $\mathcal{F}_{\mathrm{attr}}, \mathcal{F}_{\mathrm{port}}$ and semantic embeddings $\vec{\phi}_{\mathrm{attr}}, \vec{\phi}_{\mathrm{port}}$ are extracted from the two masked reference images.  
Conditioned on these features and the keypoint sequence, \dnet{} synthesizes a video of the target identity with the desired attribute through iterative denoising (\cref{eq:denoise}).

\paragraph{Multi-attribute transfer.}
Our model supports zero-shot composition of multiple attributes without additional training, by generalizing the spatial attention formulation in \cref{eq:sa_def}. Instead of using a single attribute feature, we concatenate multiple attribute feature maps along the width dimension:

\begin{equation}
\label{eq:multi}
   \bar{\mat{F}}_t^l = \mathrm{SA}\left(\mat{F}_t^l, \mat{F}_{\mathrm{port}}^l, \mathrm{concat}_{\mathrm{w}} \left(
   \mat{F}_{\mathrm{attr}}^{l,1}, 
   \mat{F}_{\mathrm{attr}}^{l,2}, 
   \cdots,
   \mat{F}_{\mathrm{attr}}^{l,N_{\mathrm{attr}}} \right) \right), 
\end{equation}

where each $\mat{F}_{\mathrm{attr}}^{l,k}$ denotes the feature map extracted from the $k$-th attribute-only image using the \arnet{}. 
To construct the final attribute-masked portrait in this setting, we also generalize the mask fusion process by taking the union of all aligned attribute masks:
\begin{equation}
\mat{M}_{\mathrm{port}}^{\mathrm{infer}} = \mat{M}_{\mathrm{port}}^{\mathrm{init}} \cup \bigcup_{k=1}^{N_{\mathrm{attr}}} \mat{M}_{\mathrm{attr}}^{\mathrm{align},k},
\end{equation}
where each \(\mat{M}_{\mathrm{attr}}^{\mathrm{align},k}\) is the aligned mask extracted from the $k$-th attribute image.
This composite mask is then used to remove all attribute regions from the portrait image before generation.
The rest of the attention computation remains unchanged, allowing the model to jointly attend to all attributes and synthesize coherent multi-attribute compositions without retraining.

\paragraph{Attribute interpolation.}
Our model enables zero-shot interpolation between two attributes of the same category (e.g., hairstyle A and B) without fine-tuning~\citep{zhang2024diffmorpher,cha2025perse}. 
Given two attribute-only images, we extract spatially attended features \(\bar{\mat{F}}_t^{\tau,l,1}\) and \(\bar{\mat{F}}_t^{\tau,l,2}\) using our spatial attention, and interpolate them as follows:

\begin{equation}
\label{eq:interp}
\bar{\mat{F}}_t^{\tau,l} = (1 - \alpha)\, \bar{\mat{F}}_t^{\tau,l,1} + \alpha\, \bar{\mat{F}}_t^{\tau,l,2},
\end{equation}

where $\alpha \in [0, 1]$ controls the interpolation ratio.
The interpolated feature \(\bar{\mat{F}}_t^{\tau,l}\) is then passed to \dnet{} for generation. This enables smooth and semantically consistent transitions between attributes.

\section{Experiments}
\label{sec:experiments}

\begin{table}[t]
  \centering
  \small
  \vspace{-3mm}
  \caption{\textbf{Quantitative Comparison.} 
  We compare our method with recent approaches that (1) synthesize portraits with transferred hairstyles, and (2) animate the synthesized portrait image. 
  }
  \label{tab:quant_image_to_video}
  \resizebox{\linewidth}{!}{ %
  \begin{tabular}{ll|ccccc|ccccc}
    \toprule
    \multicolumn{2}{l|}{} & \multicolumn{5}{c|}{\textbf{Self-Attribute Transfer}} & \multicolumn{5}{c}{\textbf{Cross-Attribute Transfer}} \\
    \cmidrule(lr){3-7} \cmidrule(lr){8-12}
    Img.Gen. & Animation & L$_1\downarrow$ & PSNR$\uparrow$ & SSIM$\uparrow$ & LPIPS$\downarrow$ & FID$\downarrow$ & mCLIP-I$\uparrow$ & mDINO$\uparrow$ & ID-Sim.$\uparrow$ & VFID$_\text{I3D}\downarrow$ & VFID$_\text{ResNeXt}\downarrow$ \\
    \midrule
    \multirow{3}{*}{PbE} 
      & LivePortrait     & 0.1059 & 16.14 & 0.5641 & 0.2859 & 40.63 & 0.8499 & 0.6407 & 0.5630 & 37.6462 & 3.3868 \\
      & X-Portrait       & 0.1180 & 15.33 & 0.5270 & 0.2978 & 59.20 & 0.8393 & 0.5916 & 0.5458 & 36.7030 & 2.9008 \\
      & MegActor-$\sum$  & 0.1268 & 14.82 & 0.4840 & 0.3157 & 62.77 & 0.8535 & 0.6266  & 0.4863 & 38.2746 & 6.2743 \\
    \midrule
    \multirow{3}{*}{HairFusion}
      & LivePortrait     & 0.1438 & 13.76 & 0.4801 & 0.3792 & 46.24 & 0.8741 & 0.6843 & 0.6502 & 30.5632 & 2.6719 \\
      & X-Portrait       & 0.1511 & 13.30 & 0.4334 & 0.3733 & 59.02 & 0.8809 & 0.6914 & 0.6520 & 30.2570 & 4.9184 \\
      & MegActor-$\sum$  & 0.1650 & 12.75 & 0.4138 & 0.4015 & 65.59 & 0.8736 & 0.6708 & 0.6044 & 30.9702 & 5.3037 \\
    \midrule
    \multirow{3}{*}{StableHair}
      & LivePortrait     & 0.1122 & 15.84 & 0.5491 & 0.3041 & 43.74 & 0.8831 & 0.7051 & 0.6564 & 29.5014 & 3.9495 \\
      & X-Portrait       & 0.1229 & 15.04 & 0.5114 & 0.3117 & 53.36 & 0.8895 & 0.7239 & 0.6443 & 28.2627 & 1.5718 \\
      & MegActor-$\sum$  & 0.1301 & 14.62 & 0.4706 & 0.3347 & 63.47 & 0.8848 & 0.7271 & 0.6130 & 30.4087 & \textbf{1.4672} \\
    \midrule
    \multirow{3}{*}{TriplaneEdit}
      & LivePortrait     & 0.1023 & 16.52 & 0.5511 & 0.2924 & 57.86 & 0.8540 & 0.6163 & 0.2776 & 32.5660 & 8.9103 \\
      & X-Portrait       & 0.1051 & 16.05 & 0.5401 & 0.2760 & 60.25 & 0.8366 & 0.6216 & 0.2944 & 30.6319 & 2.9315 \\
      & MegActor-$\sum$  & 0.1248 & 15.10 & 0.4828 & 0.3293 & 70.41 & 0.8210 & 0.5674 & 0.2770 & 32.5679 & 2.8542 \\
    \midrule
    \multicolumn{2}{l|}{\textbf{Ours}} & \textbf{0.0744} & \textbf{18.83} & \textbf{0.6527} & \textbf{0.1565} & \textbf{38.00} & \textbf{0.9043} & \textbf{0.7801} & \textbf{0.7098} & \textbf{27.1547} & 2.4052 \\
    \bottomrule
  \end{tabular}
  }
  \vspace{-3mm}
\end{table}

\begin{figure}[t]
    \centering
    \includegraphics[width=\linewidth]{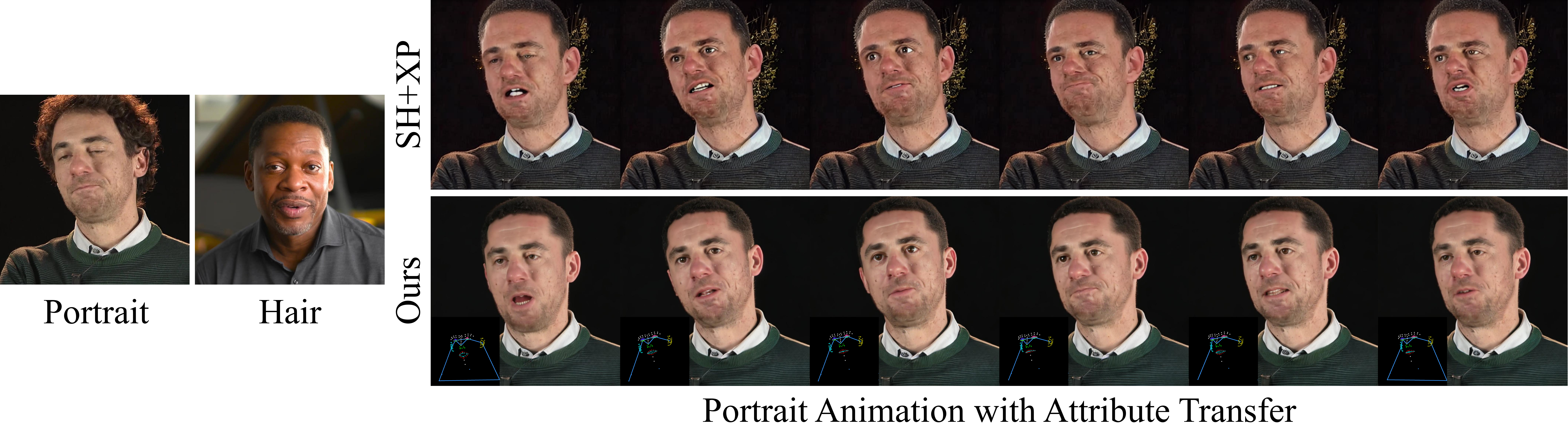}
    \caption{\textbf{Qualitative Comparison for Cross-Attribute Transfer.}
    We compare our method and the baselines that combine X-Portrait~\citep{xie2024x} with StableHair~\citep{zhang2025stablehair} in cross-identity transfer setup. We provide more results in our Supp. Mat.}
    \label{fig:qual_video_comparison}
\end{figure}

\begin{figure}[t]
    \centering
    \includegraphics[width=\linewidth]{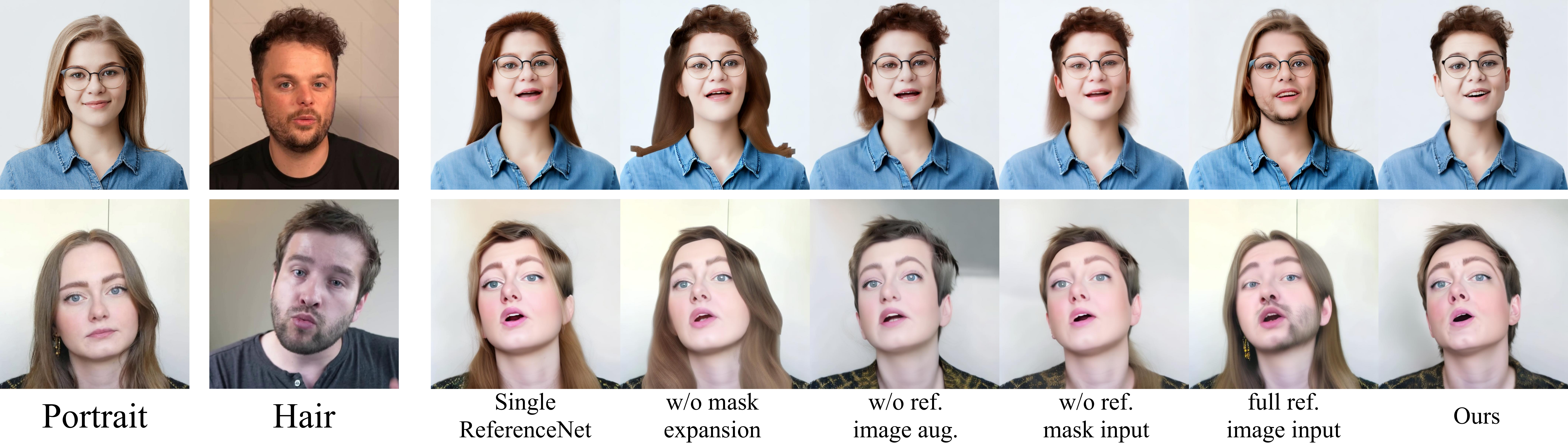}
    \caption{\textbf{Ablation Study.}
    Omitting components or altering training scheme degrades visual quality.}
    \label{fig:ablation_qualitative}
\end{figure}

\paragraph{Experimental setup.}To address the lack of ground-truth data for cross-identity attribute transfer, we design two evaluation settings: \emph{self-attribute transfer} and \emph{cross-attribute transfer}.
In \textbf{self-attribute transfer}, a single video is split into a portrait and an attribute image from different frames of the same identity, and the model reconstructs the original video.
While useful for controlled evaluation, this provides only a pseudo ground-truth and mainly reflects reconstruction ability rather than the full complexity of cross-identity transfer.
In \textbf{cross-attribute transfer}, the portrait and attribute images come from different individuals.
Without exact ground-truth, this setting instead evaluates semantic consistency, identity preservation, and temporal realism.
Together, the two settings offer a comprehensive evaluation of both low-level fidelity and high-level transfer quality.

\paragraph{Dataset.}
We train our model on CelebV-Text~\citep{yu2023celebv}, VFHQ~\citep{xie2022vfhq}, and Nersemble~\citep{kirschstein2023nersemble}, totaling 2,747 videos.  
For evaluation, we sample 200 videos for self-attribute transfer and 50 videos for cross-attribute transfer from CelebV-Text and VFHQ, ensuring diverse and unseen identities, head poses, and expressions.  
The masks for the portrait and attribute frames are generated following the procedure used in each compared method.

\paragraph{Metrics.}
For self-attribute transfer, we evaluate reconstruction fidelity using $\text{L}_{1}$, PSNR, SSIM, and LPIPS, and perceptual quality with FID~\citep{parmar2022fid}.  
For cross-attribute transfer, we measure attribute transfer quality with CLIP-I~\citep{radford2021learning, hessel2021clipscore} and DINO~\citep{caron2021emerging}, identity preservation with ArcFace~\citep{deng2019arcface}, and temporal realism with VFID~\citep{fang2024vivid} using I3D~\citep{carreira2017quo} and ResNeXt~\citep{hara2018can}.

\subsection{Comparison}
\paragraph{Baselines.}
As no prior work directly tackles portrait animation with attribute transfer from in-the-wild reference images, we construct two-stage baselines by combining image-level attribute transfer with video animation methods, resulting in 12 model combinations.
For attribute transfer (stage 1), we consider:
Paint-by-Example (PbE)~\citep{yang2023pbe}, a mask-conditioned diffusion method for reference image insertion;
HairFusion~\citep{chung2025hairfusion} and StableHair~\citep{zhang2025stablehair}, diffusion-based models for hairstyle transfer with and without masks;
and TriplaneEdit~\citep{bilecen2024reference}, a 3D-aware GAN-based face editor.
For portrait animation (stage 2), we use:
LivePortrait~\citep{guo2024liveportrait},
X-Portrait~\citep{xie2024x},
and MegActor-$\sum$~\citep{yang2025megactor}.

\paragraph{Results.}
As shown in \cref{tab:quant_image_to_video}, our method consistently outperforms all baseline combinations across both fidelity and perceptual quality metrics in self-attribute transfer.
\cref{fig:qual_video_comparison} presents a qualitative comparison against baselines using LivePortrait~\citep{guo2024liveportrait} as the animation module (stage 2).
Our method generates coherent and realistic hairstyle animations that preserve the identity and maintain consistency in spatial extent, shape, and fine details across frames. 
Please refer to our Supp. Mat. for additional qualitative comparisons with other baseline combinations.

\subsection{Ablation Study}
\begin{wraptable}{r}{0.5\textwidth} %
  \centering
  \footnotesize
  \caption{\textbf{Ablation Study.} 
  Bold indicates the best, underline the second.
  }
  \label{tab:ablation_study}
  \scriptsize
  \resizebox{\linewidth}{!}{%
  \begin{tabular}{lcccc}
    \toprule
    Variant & L$_1\downarrow$ & PSNR$\uparrow$ & SSIM$\uparrow$ & LPIPS$\downarrow$ \\
    \midrule
    single ReferenceNet     & 0.0813 & 17.95 & 0.6314 & 0.1973 \\
    w/o mask expansion      & 0.0881 & 17.16 & 0.5915 & 0.2073 \\
    w/o ref. image aug.     & 0.0900 & 16.97 & 0.5973 & 0.2248 \\
    w/o ref. mask input     & 0.0747 & 18.60 & 0.6511 & 0.1670 \\
    full ref. image input   & \textbf{0.0670} & \textbf{19.47} & \textbf{0.6698} & \textbf{0.1310} \\
    \midrule
    \textbf{Ours} & \underline{0.0744} & \underline{18.83} & \underline{0.6527} & \underline{0.1565} \\
    \bottomrule
  \end{tabular}
  }
\end{wraptable}

We evaluate the contributions of key components in our model and training strategy. \cref{tab:ablation_study} presents quantitative results, and \cref{fig:ablation_qualitative} shows corresponding qualitative comparisons.
\textbf{``single ReferenceNet''} replaces the dual-branch architecture with a shared encoder that receives the portrait and attribute images concatenated along the channel dimension, following CAT-VTON~\citep{chong2024catvton}. This setup fails to separate the roles of the two inputs, resulting in undesired blending of attribute and identity cues.
\textbf{``w/o mask expansion''} omits the attribute-aware augmentation that simulates variations in spatial extent. Without this strategy, the model tends to rely on the default shape of the portrait’s original attribute mask, making it less capable of handling diverse attribute shapes during inference.
\textbf{``w/o ref. image aug.''} disables spatial and photometric augmentations applied to the reference images during training. As a result, the model fails to accurately transfer the desired attribute with misaligned reference images.
\textbf{``w/o ref. mask input''} removes the binary mask concatenation from the inputs to the ReferenceNets. This weakens spatial localization and often leads to artifacts or residual traces of the original attribute in the output.
\textbf{``full ref. image input''} uses unmasked portrait and attribute images during training. Interestingly, this variant achieves the best quantitative scores in \cref{tab:ablation_study}, which evaluates the self-attribute transfer setting, since full images simplify the task by allowing the model to copy content more easily. However, as shown in \cref{fig:ablation_qualitative}, this model fails to disentangle identity and attribute roles, leading to visible identity leakage during cross-identity transfer.
\textbf{Ours} achieves spatially consistent, identity-preserving results, and quantitatively outperforms all other ablated variants except the full reference image variant.

\subsection{Application}
\paragraph{Multi-attribute transfer.}
\begin{figure}[t]
    \centering
    \includegraphics[width=\linewidth]{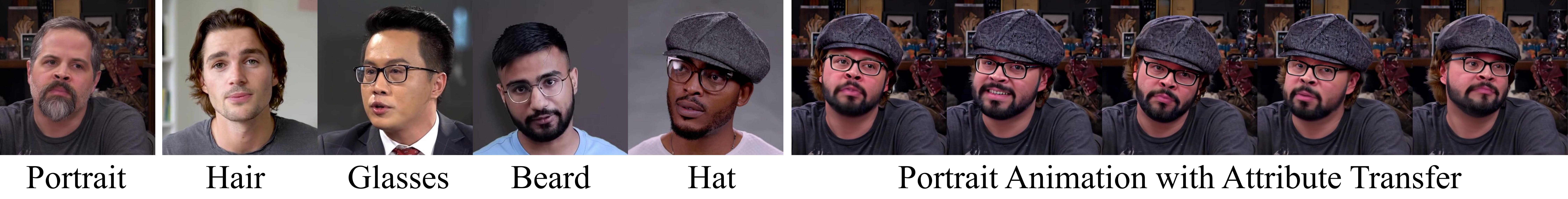}
    \caption{\textbf{Multi-Attribute Transfer.}
    Our model supports composition of multiple attributes (\emph{e.g.}, hair, eyeglasses, beard, hat) in a single forward pass without additional training.}
    \label{fig:multi_attr_composition}
\end{figure}

Our model supports the composition of multiple attributes (\emph{e.g.}, glasses, hat, hairstyle) in a single generation pass by extending the spatial attention mechanism as described in \cref{eq:multi}.
\cref{fig:multi_attr_composition} 
shows qualitative results where multiple attributes are simultaneously transferred from different reference images.
Remarkably, our model not only combines multiple attributes seamlessly but also handles interactions between overlapping regions, such as between hair and a hat.
Despite the reference images exhibiting diverse lighting conditions and spatial alignments, the model successfully integrates all attributes into the portrait image while maintaining a coherent and natural appearance.

\begin{figure}[t]
    \centering
    \includegraphics[width=\linewidth]{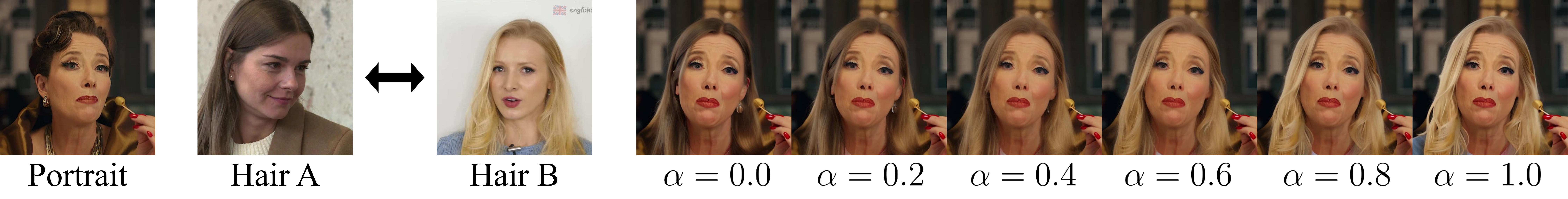}
    \caption{\textbf{Attribute Interpolation.}
    Our method generates zero-shot, single-stage portrait animations with interpolated attributes, even for rigid objects such as hats and eyeglasses. The animations interpolate naturally according to the $\alpha$ values.
    }
    \label{fig:interpolation}
\end{figure}

\paragraph{Attribute interpolation.}
Our model enables attribute interpolation by linearly blending the reference features of two attributes, as in \cref{eq:interp}.
\cref{fig:interpolation} shows the results with smooth transitions in shape and appearance.
The interpolations exhibit smooth changes in visual attributes, demonstrating that our model effectively captures semantically meaningful directions in the attribute feature space.

\section{Conclusion}

We present \modelname, a zero-shot framework for portrait animation with cross-identity attribute transfer, given a portrait image and one or more reference images specifying the target attributes. Our diffusion model, equipped with a Dual ReferenceNet, learns attribute transfer directly from uncurated portrait videos through a self-reconstruction training strategy, eliminating the need for triplet supervision. This is further enhanced by our attribute-aware mask expansion and augmentation scheme.
Moreover, \modelname{} naturally extends to multi-attribute composition and attribute interpolation within a single generation pass, without requiring any additional training.

\paragraph{Acknowledgments.}
This work was supported by NRF grant funded by the Korean government (MSIT) (No. RS-2022-NR070498), and IITP grant funded by the Korea government (MSIT) [No. RS-2024-00439854, No. 2022-0-00156, No. RS-2025-25441838, and No. RS-2021-II211343]. H. Joo is the corresponding author.

\bibliography{sections/06_reference}
\bibliographystyle{iclr2026_conference}

\end{document}